\documentclass{article}
\usepackage{spconf,amsmath,graphicx,hyperref}
\usepackage{orcidlink}
\usepackage{xcolor,booktabs}

\title{Prosody-to-Text: Predicting text from low-pass filtered speech}
\name{David Porteš\,\orcidlink{0009-0004-9279-0771} and Aleš Horák\,\orcidlink{0000-0001-6348-109X}}
\address{Natural Language Processing Centre\\
Faculty of Informatics, Masaryk University\\
Botanicka 68a, 602 00 Brno, Czech Republic\\
\texttt{{xportes,hales}@fi.muni.cz}}
\begin{document}
%
\maketitle

\begin{abstract}
While predicting prosody from text is an established task in the field, the opposite direction, predicting text that fits a given prosodic pattern, remains largely overlooked.
We find this unfortunate, because this opposite direction could lead to some very interesting use cases. Therefore, in this paper, we make the first steps in the prosody-to-text direction by investigating how much of the original sentence can be recovered from its prosodic pattern.
To this end, we fine-tune the Whisper model using only the 12 lowest Mel bins (low-pass filter with approximately 450Hz cutoff), and obtain surprisingly accurate results (WER 36\%), with 10\% of utterances being recovered perfectly, and 40\% of utterances having Word Error Rate at or below 25\%.
We also find that, given the correct prefix, the next token was predicted correctly in 79\% of cases.
Our results suggest that the relationship between low-frequency speech features and lexical content is much stronger than previously thought, and we believe that directing more attention to this topic might open the door to new applications, such as using prosody to guide text generation of modern LLMs.
\end{abstract}
\begin{keywords}
prosody, low-pass filter, Automatic Speech Recognition, Whisper, Conformer
\end{keywords}
\section{Introduction}
\label{sec:intro}

The text-prosody relationship has been extensively studied in the literature, however, the overwhelming majority of research is concerned with predicting prosody from text, due to its use in speech synthesis~\cite{galdino2025evaluation,ren2022fastspeech2fasthighquality,ogura25_interspeech}.
The opposite direction, that is, predicting text from prosody, remains largely unexplored.
While past research indicates that spoken-word recognition relies mainly on segmental features, with prosody playing a supplementary role~\cite{BuxoLugo02012026,chi2025role}, we were curious whether contemporary Automatic Speech Recognition models (ASR) could recover more than is generally thought possible from prosody alone.
If so, such prosody-to-text model could be used to recover unintelligible or corrupted speech, help transcribe dysarthric speech, or be used by researchers to study the prosody-text relationship from a new angle.
Additionally, even a partial result could be useful. For example,  one could use such model in combination with modern LLMs, in order to steer the model towards texts that fit a desired prosody on a topic of user's choice \cite{meligner,portes2025learning}, thus unlocking a new way of human-LLM interaction.

Therefore, in this paper, we investigate how much lexical content can modern speech recognition models recover from the prosodic pattern of an utterance. As a first estimate, we fine-tune the Whisper-Medium model \cite{radford2022Whisper} using only the lowest 12 Mel bins out of 80, which corresponds roughly to a low-pass filter with a 450Hz cutoff, on 750k samples from the Yodas dataset \cite{li2023yodas}. 
Low-pass filtering at approximately 400–500 Hz is a common technique to delexicalize speech while retaining prosodic information \cite{audibert23_interspeech,KNOLL2009210,10.3389/fpsyg.2015.01340}.

We obtain a surprisingly low Word Error Rate on the held out Yodas test set (WER 36\%), and on a separate Animevox~\cite{taresh18_animevox} dataset (WER 60\%).
In both datasets, there was a non-trivial percentage of sentences which were transcribed perfectly (10\%, 5\%), and, in some cases, even sentences longer than 12 words were transcribed correctly or with minimum errors.
On the Yodas test set, given the correct prefix, the model was able to predict the next token correctly 79\% of the time, and 96\% of the time within the top 10.

In order to determine how much of this result can be attributed directly to prosodic features,
we trained a Conformer-based model \cite{gulati20_interspeech} to generate artificial 12-bin Mel spectrograms based exclusively on the Fundamental frequency ($f_0$) and Energy derived from the speech.
We then continued fine-tuning our Whisper model on these generated spectrograms.
We found that, on the Yodas test set, using artificial spectrograms caused WER to increase to 69\%. However, given the correct prefix, the model was still able to predict the next token correctly 64\% of the time, and 92\% of the time in the top 10 predictions.

While we were not able to attribute our results directly to the prosodic features, we did find that the relationship between low-frequency features of speech and its lexical content is stronger than generally believed.
Our fine-tuned Whisper-Medium model, which has much fewer parameters than State of the art models, was able to provide meaningful reconstruction on a significant portion of sentences without access to most segmental features.
Our findings suggest that this might be an underexplored area with great potential, and we aim to investigate it further using additional prosodic features and larger datasets.
In order to facilitate further research in this area, we publish our models and code\footnote{\url{https://huggingface.co/collections/MU-NLPC/prosody-to-text}}.

\section{Related Work}
Predicting text from prosody is an extremely underexplored topic.
In fact, we were not able to find any works dealing with predicting text from prosody, nor low-frequency features of speech. The closest branch of research is probably the Temporal Patterns (TRAPS) \cite{BUT192490} approach  to Automatic Speech Recognition (ASR),
which proposes to conduct ASR based on long temporal patterns within different critical frequency bands, and combine bands with highest signal-to-noise ratio into the final prediction.
This implicitly includes our problem, which would be roughly equivalent to the lowest critical band. However, most of the research is dated, and we were not able to find any breakdown of accuracy by band, which we could compare to our results.

From the more recent works, the most similar is probably \cite{wolf-etal-2023-quantifying}, where its authors estimate the mutual information between prosody and text.
However, they approach the problem from the text to prosody direction, by training a LLM to predict prosody features, and they do not investigate the relationship from the prosody to text direction.

There is a number of works that treat prosody as supplementary information to an established lexical-related task, such as speech translation \cite{tsiamas-etal-2024-speech,zhou-etal-2024-prosody}, or spoken language modeling \cite{kharitonov-etal-2022-text,3737916.3742082}. However, prosody is not used as the sole input in these papers.
The only exception we are aware of is \cite{chi-etal-2025-role}, which investigates the role of prosody as supplementary information in Spoken Question Answering. Here, the authors also report the performance using only the prosodic condition, noting that the results were weaker but still very much above chance.

\begin{figure}[tb]
  \centering
  \centerline{\includegraphics[width=8.5cm]{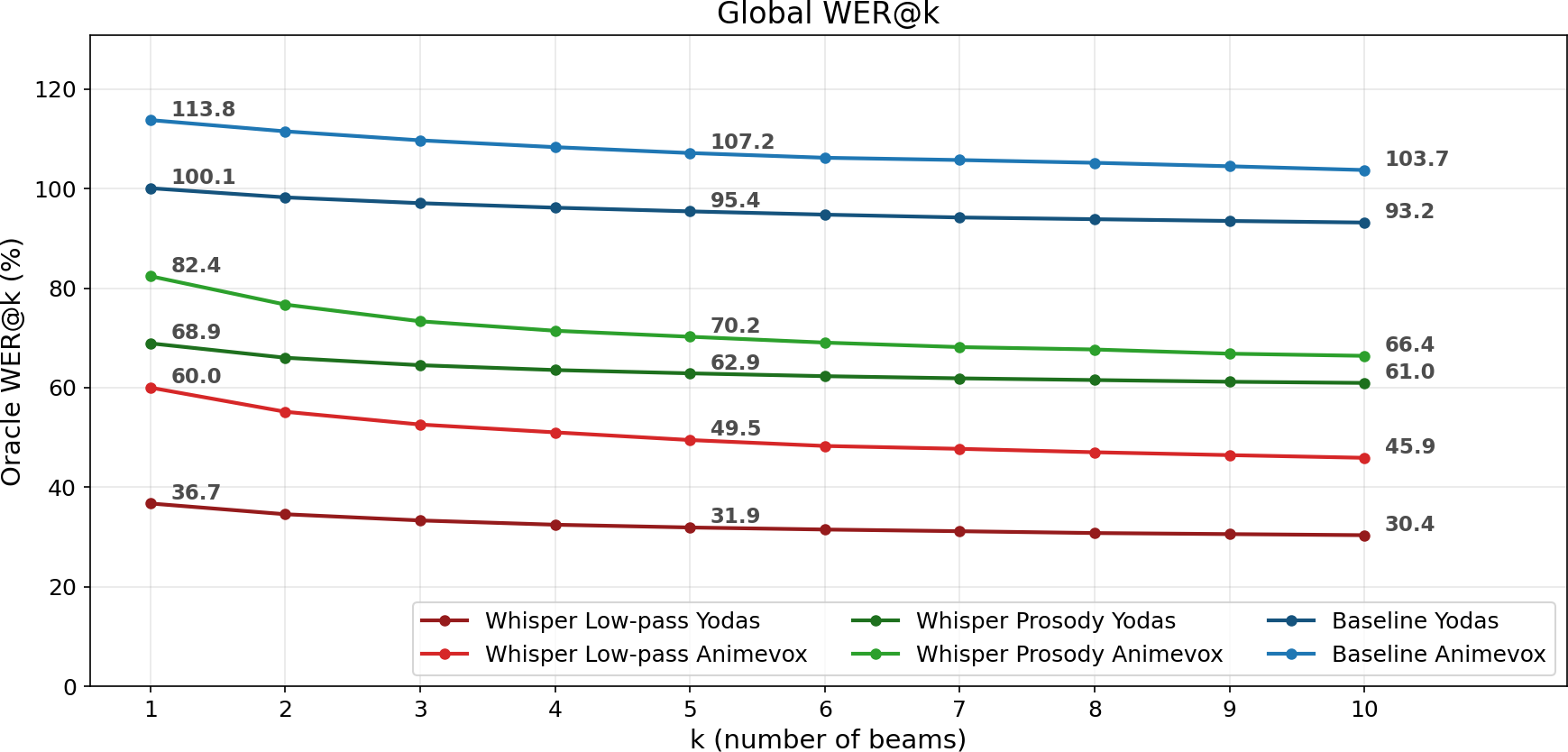}}
  \caption{The global oracle WER@k curves.}
  \label{fig:wer_global}
\end{figure}

\begin{figure}[tb]
  \centering
  \centerline{\includegraphics[width=8.5cm]{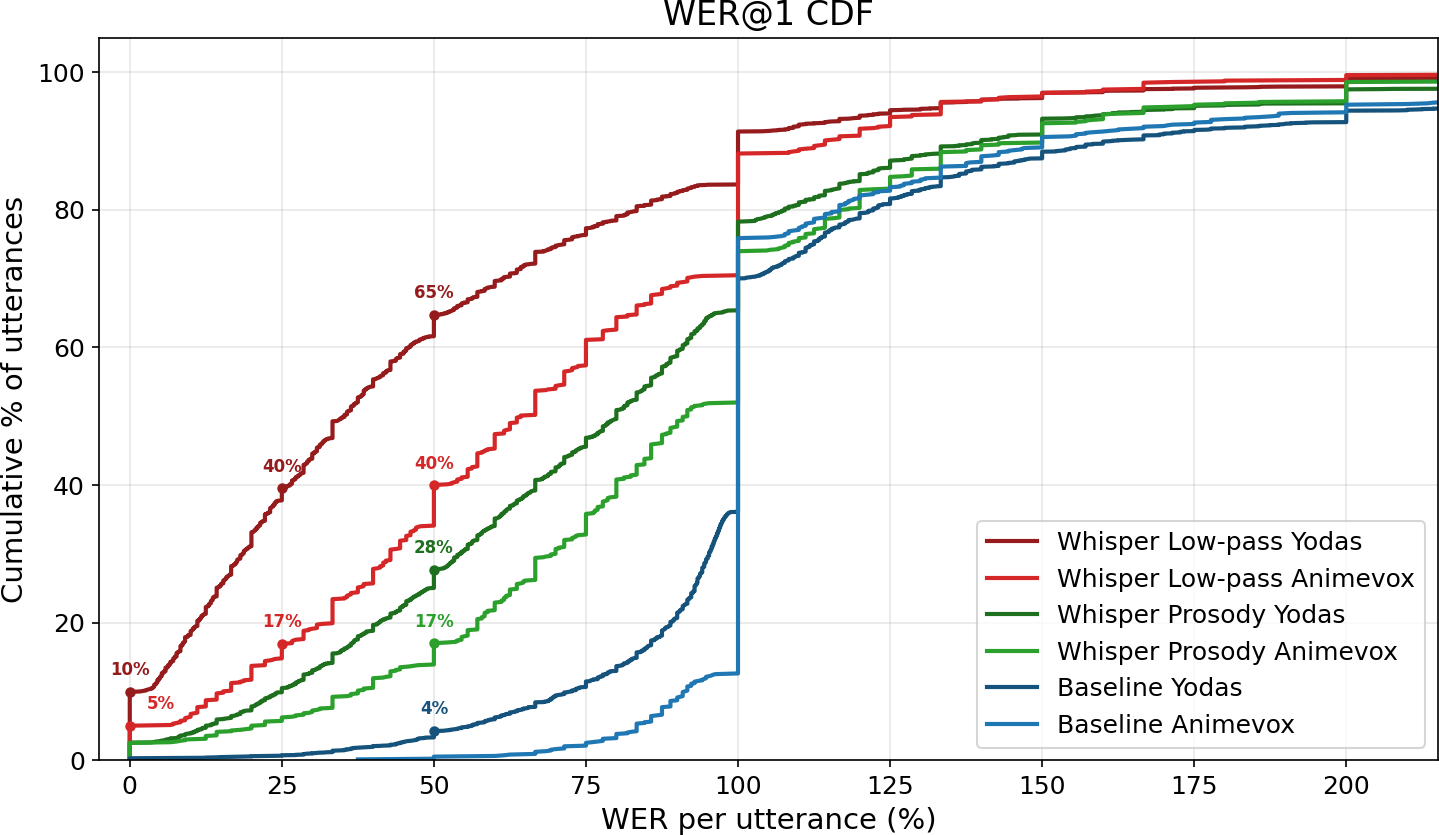}}
  \caption{Cumulative distribution over WER.}
  \label{fig:wer_dist_cdf}
\end{figure}

\section{Methods}
In this section, we describe the training of our models, and the evaluation methods used. 

\subsection{Choice of model and features}
The choice of the Whisper model was driven mainly by the fact that it is one of the most prominent ASR models, trained on a large amount of data (680k hours). While newer models could be used, it would be difficult to find a test dataset created after their release, in order to rule out any contamination.
We used the Whisper-Medium version, since we found that larger versions require much bigger batch sizes for the fine-tuning to converge.

As our representation of prosody, we used the Fundamental frequency ($f_0$) and Energy, as these are the most prominent prosodic features, often used together in speech synthesis methods~\cite{galdino2025evaluation,ren2022fastspeech2fasthighquality}. 
Fundamental frequency represents the perceived pitch of the voice, while energy reflects the loudness of the speech signal.
We decided to avoid using any text-derived features, such as duration, as these would not be available in potential real applications.
Additional spectral features are left to future work.

\subsection{Training: Whisper-Low-pass (LP)}
In order to predict text based only on the low-frequency features of speech,
we fine-tuned the Whisper-Medium model using only the lowest 12 Mel bins out of the 80 bins generated by the Whisper preprocessor. 
We did this by setting all bins above 12 to zero, in order to preserve the shape of the data. In the Whisper Mel spectrogram, this means that the highest bin we used was centered at 446Hz, tapering down towards 484Hz, where the 13th bin is centered.

We fine-tuned 750k English samples from the Yodas dataset \cite{li2023yodas}, using only samples with duration below 30 s, with a 3k validation set. We stopped fine-tuning after 1~epoch due to signs of overfitting.
Fine-tuning was conducted on a NVIDIA A100 GPU using batch size 64 with 4 gradient accumulation steps, making the effective batch size 256. The process took 12 hours.

\subsection{Training: Whisper-Prosody}
Since the Whisper model expects Mel spectrograms as input, we cannot feed it the $f_0$ and Energy prosodic features directly. For this reason, we first trained a Conformer model to generate 750k artificial 12-bin Mel spectrograms based solely on these features, using data from the Yodas dataset. The spectrograms were zero-padded to match the 80 bin shape, and they were subsequently used to further fine-tune the Whisper-Low-Pass checkpoint for an additional epoch.
Fine-tuning was conducted on a NVIDIA A100 GPU, with the same parameters and training time as in the Whisper-Low-pass case.

\subsection{Training: Conformer}
To train the Conformer model, we extracted $f_0$ using YAAPT algorithm \cite{kasi_yet_2002}, with the hop size of 5ms and frame size of 20ms on 16kHz sampled audio, resulting in $f_0$ signal sampled at 200Hz. The unvoiced frames were interpolated over using the interpolation algorithm in YAAPT. Energy was extracted using the PRAAT package \cite{boersma2001speak}, also at 200Hz.
We trained for two epochs on 750k samples from the Yodas dataset using a batch size of 256. The training took 48 hours on an NVIDIA A40 GPU.
The total number of parameters of the Conformer model was 7,645,198.

\begin{figure}[tb]
  \centering
  \centerline{\includegraphics[width=8.5cm]{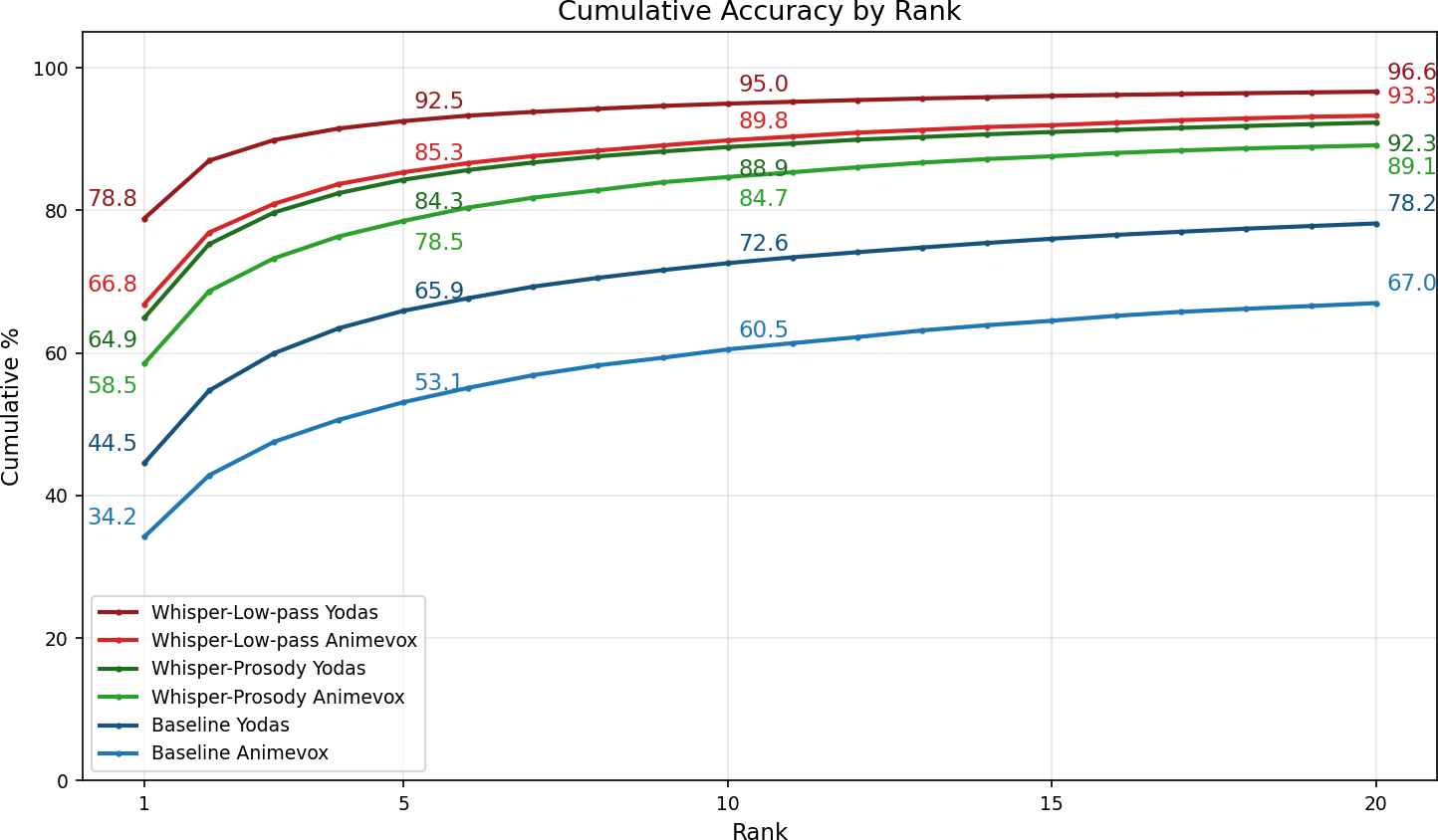}}
  \caption{Percentage of times the correct token was present in top k predictions given correct prefix.}
  \label{fig:cmp_cumulative_rank}
\end{figure}

\begin{figure}[tb]
  \centering
  \centerline{\includegraphics[width=8.5cm]{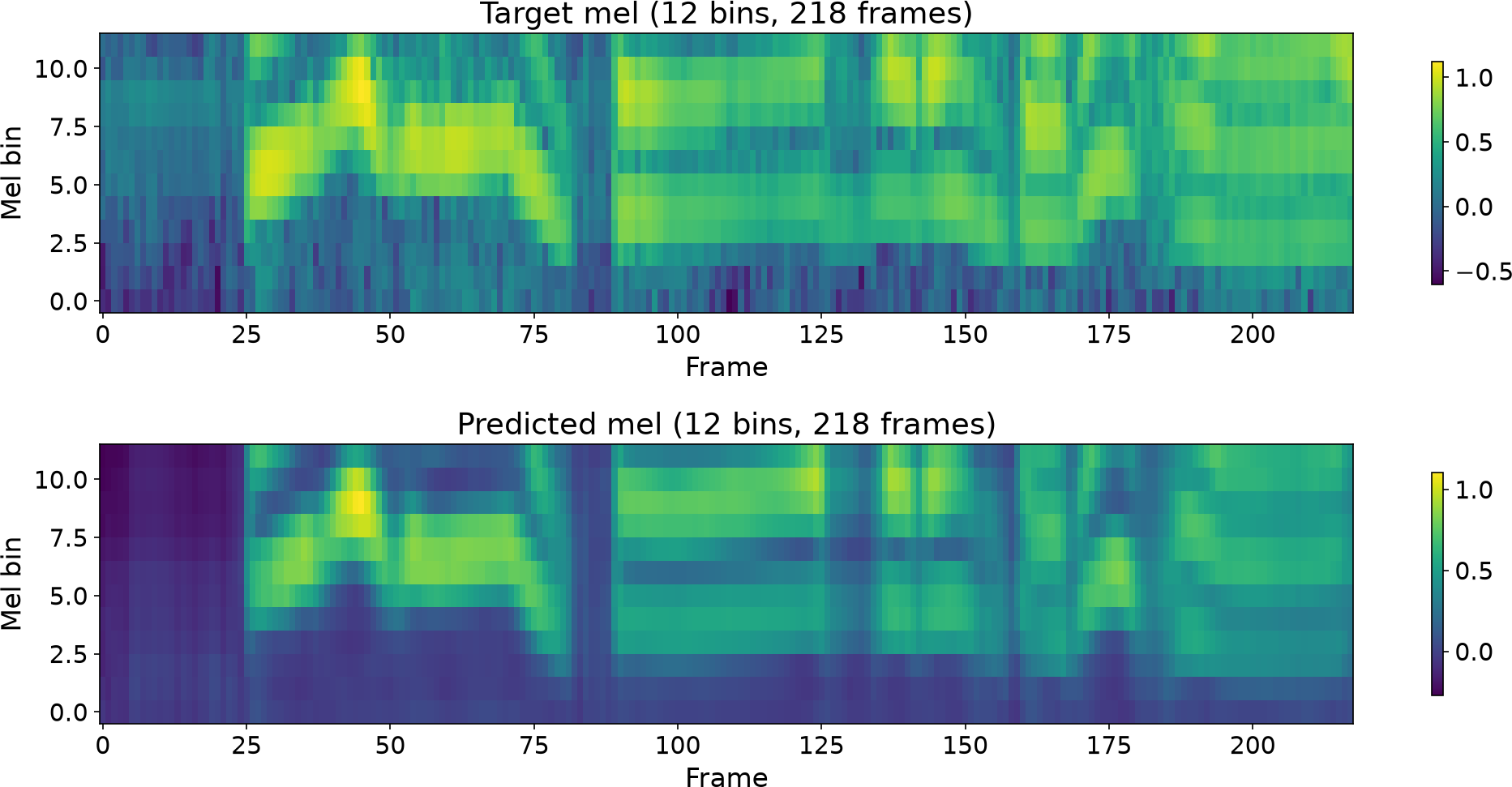}}

  \caption{Sample reconstruction of the lowest 12 bins of the Whisper Mel spectrogram by our Conformer model.}
  \label{fig:mask}

\end{figure}

\subsection{Evaluation}
We evaluated each model on 3000 held out samples of the Yodas dataset with duration under 30s. The Yodas dataset was released two years after the Whisper models, however, since the data in the dataset might be older, test set
contamination with the non-public Whisper training set cannot be ruled out.
For this reason, we also evaluated 1,000 samples from the AnimeVox dataset~\cite{taresh18_animevox}, which was made using the show \textit{Frieren: Beyond Journey's end} (2023), released a year after the Whisper model checkpoints.
For the Whisper-Prosody model, the input Mel spectrograms were generated using the Conformer model based exclusively on the prosodic features $f_0$ and Energy.
We used beam search with 10 beams in all our evaluations.
As a baseline, we use the Whisper-Large model without any fine-tuning, which was evaluated using the Low-pass condition.

\begin{figure*}[t]
\centering
\scriptsize
\def\err#1{\textcolor{red}{#1}}
\begin{tabular}{l@{\,}p{.47\textwidth}l@{\,}p{.47\textwidth}}
\multicolumn{2}{c}{\textbf{Whisper Low-pass}} &
\multicolumn{2}{c}{\textbf{Whisper Prosody}} \\
\multicolumn{2}{l}{\textbf{Yodas WER:} 37.0 \quad \textbf{ID:} {\tiny en000\_00000000\_Y65cyhwpYdo\_877\_58\_9\_80}} &
\multicolumn{2}{l}{\textbf{Yodas WER:} 68.8 \quad \textbf{ID:} {\tiny en000\_00000014\_2kON4SqVoOY\_99\_00\_4\_00}} \\
\textbf{P:} &
\textit{Unfortunately the patients knew which group they were in because \err{the} \err{other} \err{one} \err{I} \err{didn't} \err{block} \err{the} \err{rebirth,} which gave it away, \err{but} \err{she} can't discount the placebo effect.} &
\textbf{P:} &
\textit{You can now \err{come} in \err{and} \err{edit} \err{the} \err{video} \err{and} \err{there} \err{is} \err{hidden} in the \err{settings.}} \\
\textbf{R:} &
\textit{Unfortunately, the patients knew which group they were in because they were evidently getting broccoli burps, which gave it away, so you can't discount the placebo effect.} &
\textbf{R:} &
\textit{You can now send in your audition videos on the link given in the comment box.} \\[4mm]
\multicolumn{2}{l}{\textbf{Animevox WER:} 60.0 \quad \textbf{ID:} 67} &
\multicolumn{2}{l}{\textbf{Animevox WER:} 82.4 \quad \textbf{ID:} 161} \\
\textbf{P:} &
\textit{\err{Clearly} a \err{lawyer} can handle a \err{search} \err{in} \err{much} \err{longer.}} &
\textbf{P:} &
\textit{\err{Their} \err{marriage} is quite powerful, \err{such} \err{as} \err{they} \err{don't} \err{want} \err{to} \err{share} \err{with} \err{one} \err{another's} \err{child.}}\\
\textbf{R:} &
\textit{Surely a warrior can handle a fall from this height.} &
\textbf{R:} &
\textit{Though a mage is quite powerful, she could be doubly so with a warrior by her side.}\\
\end{tabular}

\caption{Sample sentences with WER near the mean for each model. \textbf{P} is prediction, \textbf{R} is reference. Red indicates errors.}
\label{fig:qualitative_reconstructions}
\end{figure*}

\section{Results and Discussion}

\begin{table}[!t]
\caption{Global Oracle WER@$k$ (\%)}
\label{tab:global_wer}
\centering
\small
\setlength{\tabcolsep}{4pt}
\begin{tabular}{@{}lcccccc@{}}
\toprule
 & \multicolumn{3}{c}{Yodas} & \multicolumn{3}{c}{AnimeVox} \\
WER & LP & Prosody & Base\textsubscript{LP} & LP & Prosody & Base\textsubscript{LP}  \\
\midrule

    @1 & \textbf{36.7} & 68.9 & 100.1 & \textbf{60.0} & 82.4 & 113.8 \\
    @2 & \textbf{34.6} & 66.0 & 98.2 & \textbf{55.2} & 76.7 & 111.5 \\
    @5 & \textbf{31.9} & 62.9 & 95.4 & \textbf{49.5} & 70.2 & 107.2 \\
    @10 & \textbf{30.4} & 61.0 & 93.2 & \textbf{45.9} & 66.4 & 103.7 \\

\bottomrule
\end{tabular}
\end{table}

\begin{table}[!t]
\caption{Next token distribution metrics}
\label{tab:analysis}
\centering
\small
\setlength{\tabcolsep}{4pt}
\begin{tabular}{@{}lcccccc@{}}
\toprule
 & \multicolumn{3}{c}{Yodas} & \multicolumn{3}{c}{AnimeVox} \\
Metric & LP & Prosody & Base\textsubscript{LP} & LP & Prosody & Base\textsubscript{LP} \\

\midrule

    Num Samples & 3000 & 3000 & 3000 & 1000 & 1000 & 1000 \\
    Top1 Pct & \textbf{78.80} & 64.90 & 44.50 & \textbf{66.80} & 58.50 & 34.20 \\
    Avg Rank & \textbf{17.33} & 39.87 & 532.44 & \textbf{81.52} & 107.14 & 959.82 \\
    Median Rank & \textbf{1} & \textbf{1} & 2 & \textbf{1} & \textbf{1} & 2 \\

    Perplexity & \textbf{2.61} & 4.91 & 29.05 & \textbf{3.78} & 5.79 & 48.54 \\

\bottomrule
\end{tabular}
\end{table}

Word Error Rates (WER) for each model-dataset combination are shown in Table~\ref{tab:global_wer}. 
Figure~\ref{fig:wer_global} visualizes the distribution of global WER values across all evaluated WER@$k$ settings, while Figure~\ref{fig:wer_dist_cdf} shows the cumulative distribution function of WER values.
In Table~\ref{tab:analysis}, we analyze several properties of the next-token distribution given the correct prefix.
Figure~\ref{fig:cmp_cumulative_rank} shows the percentage of times the correct token appeared within the top-$k$ predictions, given the correct prefix.
To illustrate average reconstruction quality, Figure~\ref{fig:qualitative_reconstructions} presents representative examples whose WER is close to the average in each model-dataset combination.

Regarding the quality of Mel spectrogram generation, 
the Mean Squared Error (MSE) between the predicted and ground-truth Mel spectrograms on the test set was $0.0322$.
Figure~\ref{fig:mask} shows a representative reconstruction.

We find that, given the correct prefix, the Whisper-Low-pass model predicted the correct next token approximately 79\% of the time on the Yodas dataset and 67\% on the AnimeVox dataset (Table~\ref{tab:analysis}).
While this was insufficient for reliable reconstruction,
the WER was very unstable, and a nontrivial portion of sentences were even reconstructed with 0\% WER (10\%/5\%).
There was also a substantial gap between the two datasets, which we suspect to be due to the domain mismatch between the fine-tuning dataset consisting of Youtube videos (Yodas) and Animevox, containing mostly animated voice acting in a fantasy setting.

We were not able to attribute these results purely to the prosodic features of $f_0$ and Energy, as WER increased substantially when using the Whisper-Prosody pipeline.
This may indicate that prosodic information was only partially responsible for the performance of the Whisper-Low-pass model, or that additional prosodic features are required. It may also have been caused by insufficient training data, as Whisper-Prosody relies on a more complex two-stage pipeline involving the Conformer model.
Even so, the correct next-token rate of the Whisper-Prosody-based model remained above 58\% on both datasets, suggesting the presence of a non-trivial relationship between $f_0$, Energy and lexical content.

Our results support the base hypothesis behind TRAPS~\cite{BUT192490} that individual frequency bands carry large portions meaning redundantly, at least in the lowest critical band.
In this context, it would be interesting to see whether current state of the art ASR models draw on the low-frequency features to aid their predictions in cases where the signal in other bands is corrupted.

When compared to the base Whisper models, we find that this is not a capability that the Whisper model possessed natively, as our fine-tuned Whisper-Medium  model's WER (36 / 60\%) is much lower than the WER of even the Whisper-Large model (WER 100 / 113\%), across both datasets.

\section{Limitations}
The main limitation of our study is that, in case of the Yodas dataset, it is not possible to rule out test set contamination with the original Whisper training set, since it is not publicly available.
We have also extracted the test set in an interleaved manner with the training set, meaning that the domain is mostly the same across both sets, and using a disjoint set of the Yodas dataset might yield worse results.
However, the results on AnimeVox are guaranteed to be free from contamination, since the data comes from a TV show recorded one year after the Whisper checkpoints last modification date on the Huggingface website.
The AnimeVox dataset is not an established dataset in the field, however, we were hard-pressed to find any other datasets with samples that are guaranteed to have been created after the Whisper model cutoff.
The data used from the AnimeVox dataset also encompass only a single female speaker.

\section{Conclusion}
Our results reveal a surprisingly strong relationship between low-frequency features of speech and its lexical content. On the AnimeVox dataset, using only the lowest 12 Mel bins (450Hz), our fine-tuned Whisper-Medium model was able to correctly transcribe 5\% of utterances,
with ~17\% of utterances having Word Error Rate at or below 25\%. Our results were even better on the Yodas test set, with 10\% fully correct utterances, and 40\% utterances at or below 25\% WER. However, in this case,
it is not possible to rule out test set contamination with the
original Whisper training set.
While the prosody-only results were weaker, there was still indication of a nontrivial prosody-text relationship, as our model's next token predictions were correct in more than 58 percent of cases, given the correct prefix.
All our results were obtained using the Whisper-Medium model, fine-tuned on very small data compared to today's industrial standard, suggesting that there is still potential to improve the performance further, which could lead to new original use cases and applications.
As future work, we plan to add additional prosodic features to see whether the performance on low-pass filtered data can be matched.

\vfill\pagebreak
\subsection*{Acknowledgments}
 
Computational resources were provided by the e-INFRA CZ project (ID:90254), supported by the Ministry of Education, Youth and Sports of the Czech Republic.




\bibliographystyle{IEEEbib}
\bibliography{custom}
\end{document}